\documentclass[journal]{IEEEtran}
\usepackage{cite}

\ifCLASSINFOpdf
\else
\fi
\usepackage{array}

 \usepackage[caption=false,font=footnotesize]{subfig}

\usepackage{times}
\usepackage{epsfig}
\usepackage{graphicx}
\usepackage{amsmath}
\usepackage{amssymb}
\usepackage{epstopdf}
\usepackage{multirow}
\usepackage{CJK}
\newcommand{\tabincell}[2]{\begin{tabular}{@{}#1@{}}#2
\end{tabular}}
\usepackage{subfig}
\usepackage{diagbox}

\usepackage{color}

\begin{document}
%
\title{Action Recognition Based on Joint Trajectory Maps with Convolutional Neural Networks}
%
%
%


\author{Pichao Wang,~\IEEEmembership{Student Member,~IEEE,}
        Wanqing Li,~\IEEEmembership{Senior Member,~IEEE,}\\
        Chuankun Li, and 
        Yonghong Hou,~\IEEEmembership{Member,~IEEE,}
 \thanks{Manuscript received XXX; revised XXX. This work was supported by the National Natural Science Foundation of China (grant 61571325) and Key Projects in the Tianjin Science \& Technology Pillar Program (grant 15ZCZD GX001900). (Corresponding author: Yonghong Hou)}
 \thanks{P. Wang and W. Li  are with the 
Advanced Multimedia Research Lab, University of Wollongong, Wollongong, 
Australia.
(e-mail: pw212@uowmail.edu.au; wanqing@uow.edu.au).}
\thanks{C. Li and Y. Hou are with the School of Electronic Information Engineering, Tianjin University, Tianjin, China. (e-mail:chuankunli@tju.edu.cn;houroy@tju.edu.cn).}}

%
%

\markboth{Submitted to IEEE TRANSACTIONS ON CYBERNETICS,~Vol.~X, No.~X, X~2016 (WILL BE INSERTED BY THE EDITOR)}%
{Shell \MakeLowercase{\textit{et al.}}:Action Recognition Based on Joint Trajectory Maps with Convolutional Neural Networks}
%



\maketitle

\begin{abstract}
Convolutional Neural Networks (ConvNets) have recently shown promising performance in many computer vision tasks, especially image-based recognition. How to effectively apply ConvNets to sequence-based data is still an open problem. This paper proposes an effective yet simple method to represent spatio-temporal information carried in $3D$ skeleton sequences into three $2D$ images by encoding the joint trajectories and their dynamics into color distribution in the images, referred to as Joint Trajectory Maps (JTM), and adopts ConvNets to learn the discriminative features for human action recognition. Such an image-based representation enables us to fine-tune existing ConvNets models for the classification of skeleton sequences without training the networks afresh. The three JTMs are generated in three orthogonal planes and provide complimentary information to each other. The final recognition is further improved through multiply score fusion of the three JTMs. The proposed method was evaluated on four public benchmark datasets, the large NTU RGB+D Dataset, MSRC-12 Kinect Gesture Dataset (MSRC-12), G3D Dataset and UTD Multimodal Human Action Dataset (UTD-MHAD) and achieved the state-of-the-art results.
\end{abstract}


%
\IEEEpeerreviewmaketitle

\section{Introduction}
%
%
%
%

\IEEEPARstart{H}{uman} action recognition is an important problem in computer vision due to its wide applications in video surveillance, human computer interfaces, robotics, etc. Despite significant research efforts over the past few decades~\cite{darrell1993space,campbell1995recognition,bobick1996appearance,bobick2001recognition,yilmaz2005actions,
laptev2005space,Li2008,wang2013action,liu2013learning,wang2013dense,peng2014action,shao2014spatio,kantorov2014efficient,ni2015motion,wang2015action,liu2016learning,
bilen2016dynamic,Shao2016,peng2016bag,Fernando2016b,wang2016temporal}, accurate recognition of human actions from RGB video sequences is still an unsolved problem. With the advent of easy-to-use and low-cost depth sensors such as {MS} Kinect sensors,  human action recognition from RGB-D (Red, Green, Blue and Depth) data has attracted increasing attention and many applications have been developed~\cite{han2013enhanced} in recent years, due to the advantages of depth information over conventional RGB video, e.g. being insensitive to illumination changes and reliable to estimate body silhouette and skeleton~\cite{Shotton2011}. Since the first work~\cite{li2010action} reported in 2010, many methods~\cite{Yang2012a,wang2012mining,Oreifej2013,yangsuper,lurange} have been proposed using specifically hand-crafted feature descriptors extracted from depth. As the extraction of skeletons from depth maps~\cite{Shotton2011} has become increasingly robust, more and more hand-designed skeleton features~\cite{Yang2012,zanfir2013moving,Gowayyed2013_HOD,pichao2014,vemulapalli2014human,xia2012view,shao2013new,chaudhry2013bio,
ohn2013joint,devanne20153,vemulapalli2016r3dg,vemulapallirolling,yang2016latent}  have been devised to capture spatial configuration, and Dynamic Time Warpings (DTWs), Fourier Temporal Pyramid (FTP) or Hidden Markov Models (HMMs) are employed to model temporal information. However, these hand-crafted features are always shallow and dataset-dependent.
Recently, Recurrent Neural Networks (RNNs)~\cite{du2015hierarchical,veeriah2015differential,zhu2015co,shahroudy2016ntu,liu2016spatio} have also been adopted for action recognition from skeleton data. RNNs tend to overemphasize the temporal information especially when the training data is not sufficient, leading to overfitting. Up to date, it remains unclear how skeleton sequences could be effectively represented and fed to deep neural networks for recognition. For example, one can conventionally consider a skeleton sequence as a set of individual frames with some form of temporal smoothness, or as a subspace
of poses or pose features, or as the output of a neural network encoder. Which one among these and other possibilities would result in the best representation in the context of action recognition is not well understood.
\begin{figure}[t]
\begin{center}
{\includegraphics[height = 50mm, width = 70mm]{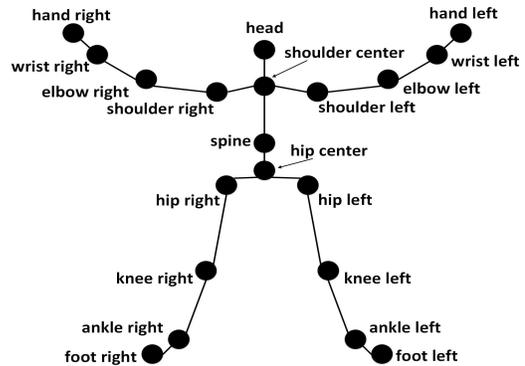}}
\end{center}
\caption{The joint configuration for Kinect V1 skeleton.  }
\label{skl}
\end{figure}

\begin{figure*}[!ht]
\begin{center}
{\includegraphics[height = 65mm, width = 180mm]{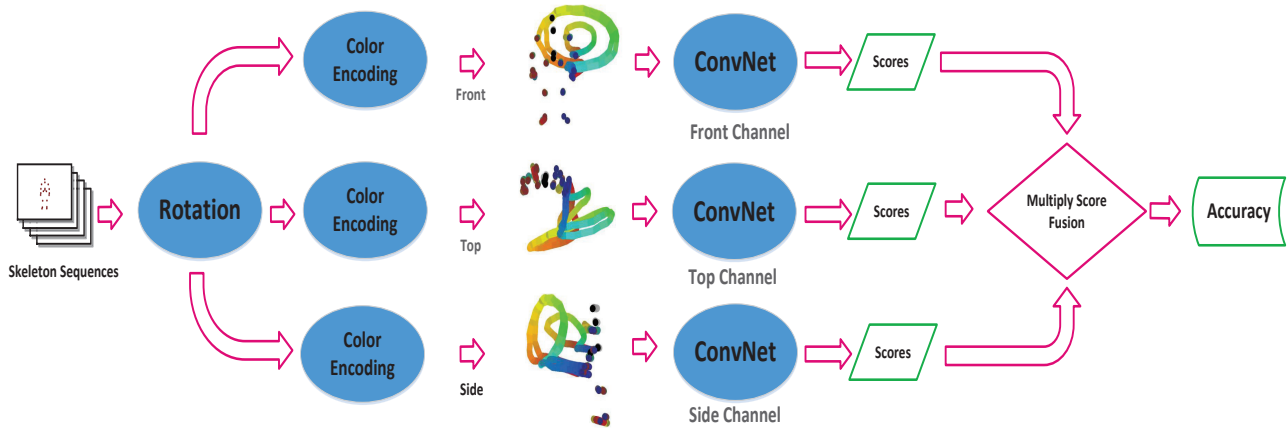}}
\end{center}
\caption{The framework of the proposed method.}
\label{fig:framework}
\end{figure*}

In this paper, we present an effective yet simple method that represent both spatial configuration and dynamics of joint trajectories into three texture images through color encoding, referred to as Joint Trajectory Maps (JTMs), as the input of ConvNets for action recognition. Such image-based representation enables us to fine-tune existing ConvNets models trained on ImageNet for classification of skeleton sequences without training the whole deep networks afresh. The three JTMs are complimentary to each other, and the final recognition accuracy is improved largely by a late score fusion method.  One of the challenges in action recognition is how to properly model and use the spatio-temporal information. The commonly used bag-of-words model often ignores spatial information. On the other hand, HMMs or RNNs based methods are likely to overstress the temporal information. The proposed method addresses this challenge in a novel way by encoding as much the spatio-temporal information as possible (without a need to decide which one is important and how important it is) into images, and employing ConvNets to learn the discriminative one. Consequently, the proposed method outperformed the start-of-the-art methods on popular benchmark datasets. 

The main contributions  of this paper include:
\begin{itemize}
\item A compact, effective yet simple image-based representation is proposed to represent the spatio-temporal information carried in the $3D$ skeleton sequences into three $2D$ images by encoding the dynamics of joint trajectories into three complementary Joint Trajectory Maps.  

\item To overcome the drawbacks of ConvNets not being rotation-invariant, and to make the proposed method suitable for cross-view action recognition,  it is proposed to rotate the skeleton data to not only mimic the multiple views but also to augment data effectively for training. 

\item The proposed method was evaluated  on four popular public benchmark datasets, namely, the large NTU RGB+D Dataset~\cite{shahroudy2016ntu}, MSRC-12 Kinect Gesture Dataset (MSRC-12)~\cite{Fothergill2011}, G3D Dataset~\cite{bloom2012g3d} and UTD Multimodal Human Action Dataset (UTD-MHAD)~\cite{chenchen2015b}, and achieved the state-of-the-art recognition results. 
\end{itemize}

This paper is an extension of the works presented in~\cite{pichao2016,Hou2016}. Unlike~\cite{pichao2016,Hou2016} where skeletons are assumed to have been sufficiently sampled and discrete joints are drawn onto images using a pen whose size is properly set, this paper employs joint trajectories and proposes to rotate skeletons to mimic multiple views for cross-view action recognition and data augmentation. In addition, this paper adopts multiply score fusion to improve the final recognition accuracy. Extensive experiments and detailed analysis are also presented in this paper. The rest of this paper is organized as follows. An overview of related works is given in Section~\ref{relatedwork}. Details of the proposed method are described in Section~\ref{proposedmethod}. Evaluation of the proposed method on four datasets and analysis of the results are reported in Section~\ref{experiments}. Section~\ref{conclusion} concludes the paper with remarks.

\section{Related Work}\label{relatedwork}
An extensive review on RGB-D based action recognition is beyond the scope of this paper. Readers are referred to~\cite{aggarwal2014human,presti20163d,zhang2016rgb} for a comprehensive survey. In this section, the work related to the proposed method is briefly reviewed, including skeleton-based {3D} action representation and deep learning based action recognition.

\subsection{Skeleton-Based 3D Action Representation}
Skeleton based 3D action representation can be generally divided into three categories~\cite{presti20163d}: joints, groups of joints , and joint dynamics. Joint representation captures the correlation of the body joints by extracting spatial descriptor~\cite{Yang2012,ellis2013exploring,Hussein2013,kerola2014spectral,wu2014leveraging}, geometric descriptor~\cite{evangelidis2014skeletal,pichao2014,vemulapalli2014human,devanne20153,vemulapalli2016r3dg,vemulapallirolling} or key poses~\cite{xia2012view,ofli2014sequence,barnachon2014ongoing,lillo2014discriminative}. The groups of joints aim to detect the discriminative subsets of joints to differentiate actions. Methods such as~\cite{ofli2014sequence,wang2013approach,wei2013concurrent,eweiwi2014efficient,pichao2014,Shahroudy2016} focus on mining the subsets of most discriminative joints or consider the correlation of predefined subsets of joints. 

Joint dynamics focuses on modeling the dynamics of either subsets or all joints of a skeleton. In~\cite{Gowayyed2013_HOD} 3D trajectories of joints are projected into three 2D trajectories, and histogram of oriented displacement is calculated to describe the three 2D trajectories, with each displacement in the trajectory voting its length in the histogram of orientation angles.  Chaudhry et al.~\cite{chaudhry2013bio} divided the fully body skeleton into several body parts represented by joints, including the upper body, lower body, left/right arms and left/right legs. A shape context feature is computed by considering the directions of a set of equidistant points sub-sampled over the segments of each body part.  A skeleton sequence is finally represented as a set of time series of features such as position, tangent and shape context feature. These time series are further divided into several temporal scales, and each individual feature series is modeled using a linear dynamic system. The estimated parameters of all series are used to describe the dynamics of the skeleton sequence. In~\cite{zanfir2013moving} a skeleton sequence is modeled as a continuous and differentiable function of the body joint locations over time. The local 3D body pose is characterized by the current joint locations and differential properties like speed and acceleration of the joints.
Slama et al.~\cite{slama2015accurate} represented each action sequence as a linear dynamic system that produces 3D joint trajectories. Autoregressive moving average model was adopted to represent the dynamics by means of observability matrix which embeds the parameters of the model. In~\cite{lehrmann2014efficient} the dynamic forest model was proposed and a set of autoregressive trees was adopted. Each node in the  probabilistic autoregressive tree stores a multivariate normal distribution with a fixed covariance matrix, and the set of Gaussian posteriors estimated by the forest are used to calculate the forest posterior. Shao et al.~\cite{shao2015integral} proposed to use a class of integral invariants to describe motion trajectories by calculating the line integral of a class of kernel functions at multiple scales along the motion trajectory. In~\cite{devanne20153} the authors represented the 3D coordinates of joints and their changes over time as a trajectory in the Riemannian manifold, and the action recognition is formulated as the problem of computing the similarity between the shape of trajectories. In this paper, we propose to use color to encode the dynamics of trajectories, and model the spatial-temporal information carried in a skeleton sequence through shape and textures. ConvNets are used to learn deep hierarchy features. 

\subsection{Deep Leaning Based Action Recognition}
The exiting deep learning approaches to action recognition can be generally divided into four categories based on how an input sequence is represented and fed to a deep neural network. The first category views a video either as a set of still images~\cite{yue2015beyond} or as a short and smooth transition between similar frames~\cite{simonyan2014two}, and each color channel of the images is fed to one channel of a ConvNet. Although suboptimal, considering the video as a bag of static frames gives reasonable results. The second category is to represent a video as a volume and extends ConvNets to a third, temporal dimension~\cite{ji20133d,tran2015learning} replacing 2D filters with 3D ones. So far, this approach has produced little benefits, probably due to the lack of annotated training data. The third category is to treat a video as a sequence of images and feed the sequence to a RNN~\cite{donahue2015long,du2015hierarchical,veeriah2015differential,zhu2015co,sharma2015action,shahroudy2016ntu}. A RNN is typically considered as memory cells, which are sensitive to both short as well as long term patterns. It parses the video frames sequentially and encodes the frame-level information in their memory. However, using RNNs has not given an improvement over temporal pooling of convolutional features~\cite{yue2015beyond} or over hand-crafted features. The last category is to represent a video in one or multiple compact images and adopt available trained ConvNet architectures for fine-tuning~\cite{pichao2015,pichaoTHMS,pichao2016,bilen2016dynamic,Hou2016}. This approach has achieved state-of-the-art results  on many RGB and depth/skeleton datasets. The proposed method falls into this category. 

\section{The Proposed Method}\label{proposedmethod}

The proposed method consists of four major components, as illustrated in Fig.~\ref{fig:framework}, rotation to mimic the multiple views, construction of three JTMs as the input of the ConvNets in three orthogonal planes from skeleton sequences, training the three ConvNets to learn discriminative features, and multiply score fusion for final classification. In the following sections, the four components are detailed.

\subsection{Rotation}
A skeleton is often represented by a set of joints in 3D space with respect to the real-world coordinate system centered at the optical central of the RGB-D camera. By rotating the skeleton data, it can 1) mimic multi-views for cross-view action recognition; 2) enlarge the data for training and overcome the drawback of ConvNets usually being not view-invariant.  

The rotation was performed with a fixed step of $15^{\circ}$ along the polar angle $\theta$ and azimuthal angle $\psi$, in the range of  $[0^{\circ}, 45^{\circ}]$ for $\theta$ and $[-45^{\circ}, 45^{\circ}]$ for $\psi$. The ranges of $\theta$ and $\psi$ would cover the possible views considering that the JTMs are generated by projecting the trajectories onto the three orthogonal planes as detailed below.

Let $\mathbf Tr_{y}$ be the transform around $y$ axis (right-handed coordinate system) and $\mathbf Tr_{x}$ be the transform around $x$ axis. The coordinates $(x_{r}, y_{r}, z_{r})$  of a joint at $(x, y, z)$  after rotation can be expressed as 
\begin{equation}\label{eq2}
\mathbf [x_{r}, y_{r}, z_{r}, 1]^ \mathrm{ T }  = \mathbf Tr_{y}\mathbf Tr_{x}
\begin{bmatrix}
 x, y, z, 1 \\     
\end{bmatrix}^ \mathrm{ T }
\end{equation}

where
\begin{equation}\label{eq2-2}
\mathbf Tr_{y} = 
\begin{bmatrix}
 R_{y}(\psi) & T_{y}(\psi)\\     
 \textbf{0} & 1    
\end{bmatrix}; 
\mathbf Tr_{x} = 
\begin{bmatrix}
 R_{x}(\theta) & T_{x}(\theta)\\     
 \textbf{0} & 1 
\end{bmatrix}, \\
\end{equation}
and\\
\begin{footnotesize}
$\mathbf R_{y}(\psi) = 
\begin{bmatrix}
 1 & 0 & 0 \\     
 0 & \cos(\psi) & -\sin(\psi)\\ 
 0 & \sin(\psi) & \cos(\psi) \\ 
\end{bmatrix}$
$\mathbf T_{y}(\psi) = 
\begin{bmatrix}
 0\\
z\cdot \sin(\psi)\\     
z\cdot (1 - \cos(\psi))\\     
\end{bmatrix};$
$\mathbf R_{x}(\theta) = 
\begin{bmatrix}
 cos(\theta) & 0 & \sin(\theta) \\     
 0 & 1 & 0 \\ 
 -\sin(\theta) & 0 & \cos(\theta) \\ 
\end{bmatrix}$
$\mathbf T_{x}(\theta) = 
\begin{bmatrix}
 -z\cdot \sin(\theta)\\
 0\\     
z\cdot (1 - \cos(\theta))\\     
\end{bmatrix}. $
\end{footnotesize}

\subsection{Construction of JTMs}
We argue that an effective JTM should have the following properties to keep the spatial-temporal information of an action:
\begin{itemize}
\item The joints or group of joints should be distinct in the JTM such that the spatial information of the joints is well reserved.
\item The JTM should encode effectively the temporal evolution, i.e. trajectories of the joints, including the direction and speed of joint motions.
\item The JTM should be able to encode the difference in motion among the different joints or parts of the body to reflect how the joints are synchronized during the action.
\end{itemize}

Specifically, a JTM can be recursively defined  as follows
\begin{equation}
JTM_{i} = JTM_{i -1} + f(i),
\label{eq:spectrum}
\end{equation}
where $f(i)$ is a function encoding the spatial-temporal information at frame or time-stamp $i$. Since a JTM is accumulated over the period of an action, $f(i)$ has to be carefully defined such that the JTM for an action sample has the required properties discussed above and the accumulation over time has little adverse impact on the spatial-temporal information that has already been encoded in the JTM. 
This paper proposes to use hue, saturation and brightness to encode the spatial-temporal motion patterns.

\begin{figure}[!ht]
\begin{center}
{\includegraphics[height = 40mm, width = 85mm]{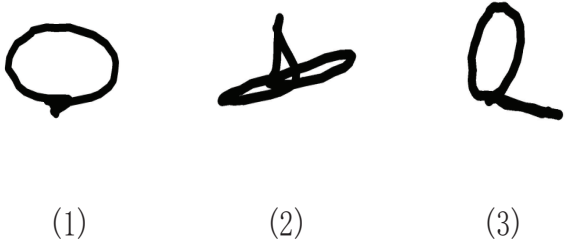}}
\end{center}
\caption{The trajectories projected onto three Cartesian planes for action ``right hand draw circle (clockwise)" in {UTD-MHAD~\protect\cite{chenchen2015b}}: (1) front plane; (2) top plane; (3) side plane.}
\label{fig1}
\end{figure}

\subsubsection{Joint Trajectory Maps}
Assume an action $H$ has $n$ frames of skeletons and each skeleton consists of $m$ joints. The skeleton sequence is denoted as $H=\{F_{1},F_{2},...,F_{n}\}$, where $F_{i} = \{P_{1}^{i},P_{2}^{i},...,P_{m}^{i}\}$ is a vector of joint coordinates of frame $i$, and $P_{j}^{i}$ is the $3D$ coordinates of the $j^{th}$ joint in frame $i$. The skeleton trajectory $T$ for an action of $n$ frames consists of the trajectories of all joints and is defined as:
\begin{equation}
T = \{T_1, T_2, \cdots,T_i,\cdots,T_{n-1}\},
\label{eq:actionrepre}
\end{equation}
where $T_{i} = \{t_{1}^{i},t_{2}^{i},...,t_{m}^{i}\} = F_{i+1} - F_{i}$, and the $k^{th}$ joint trajectory is $t_{k}^{i} = P_{k}^{i+1} - P_{k}^{i}$. A simple form of function $f(i)$ would be $T_{i}$, that is,
\begin{equation}
f(i) = {T_i} = \{ {t_1^i,t_2^i,...,t_m^i} \}.
\end{equation}
 
The skeleton trajectory is projected to three orthogonal planes, i.e. three Cartesian planes of the real world coordinates of the camera, to form three JTMs. Fig.~\ref{fig1} shows the three projected trajectories of the right hand joint for action ``right hand draw circle (clockwise)" in the UTD-MHAD dataset. It can be seen that the spatial information of this joint over the period of the action is well represented in the JTMs but the direction of the motion is lost.

\subsubsection{Encoding Joint Motion Direction} 
\begin{figure}[!ht]
\begin{center}
{\includegraphics[height = 50mm, width = 85mm]{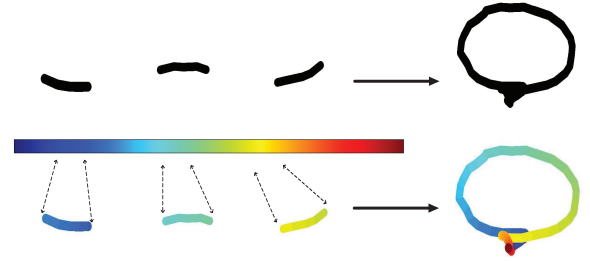}}
\end{center}
\caption{An example of colored coded joint trajectory with different colors reflecting the temporal order.}
\label{fig2}
\end{figure}

To capture the motion direction in the JTM, it is proposed to use hue to ``color" the joint trajectories over the action period. Different colormaps may be chosen. In this paper, the jet colormap, ranging from blue to red, and passing through the colors cyan, yellow, and orange, is adopted. Let the color of a joint trajectory be $C$, and the length of the trajectory be $L$, and $C_{l}, l \in (0, L)$ be the color at position $l$ of a trajectory. For the $q^{th}$ trajectory $T_{q}$ from $1$ to $n-1$, a color $C_{l}$, where $l = \frac{q}{n-1}\times L$ is assigned to location $l$ of the joint trajectory, making the entire trajectory colored over the period of the sequence as illustrated in Fig.~\ref{fig2}. Herein, a trajectory with color is denoted as $C\_t_k^i$ and the function $f(i)$ becomes:
\begin{equation}
f(i) = \{ {C\_t_1^i,C\_t_2^i,...,C\_t_m^i} \}.
\end{equation}
Fig.~\ref{fig3} shows the front JTM of action ``right hand draw circle (clockwise)'' in the UTD-MHAD~\cite{chenchen2015b} dataset. Sub-figure (1) is joint trajectories and sub-figure (2) is the trajectories with motion direction being encoded with hue. The color variations along the trajectories represent the motion direction.

\subsubsection{Encoding Body Parts}
Many actions, especially complex actions, often involve multiple body parts and these body parts move in a coordinating manner. It is important to capture such coordination in the JTMs. To distinguish different body parts, multiple colormaps are employed. Body parts can be defined at different levels of granularity. For example, each joint can be considered independently as a ``part" and is assigned to one colormap, or several groups of joints can be defined and all joints in each group are assigned to the same colormap and colormaps are chosen randomly to each group. Since arms and legs often move more than other body parts, a body is divided into three parts in this paper. According to the joint configuration for Kinect V1 skeleton as shown in Fig.~\ref{skl},  the left body part consists of left shoulder, left elbow, left wrist, left hand, left hip, left knee, left ankle and left foot, the right body part consists of right shoulder, right elbow, right wrist, right hand, right hip, right knee, right ankle and right foot and the middle part consists of head, neck, torso and hip center. The three parts are assigned to three colormaps ($C1, C2, C3$) respectively, where $C1$ is the same as $C$, i.e. the jet colormap, $C2$ is a colormap with reversely-ordered  colors of $C1$, and $C3$ is a gray-scale map ranging from light gray to black. Let the trajectory encoded by multiple colormaps be $MC\_t_k^i$.  Function $f(i)$ can be expressed as:
\begin{equation}
f(i) = \{ {MC\_t_1^i,MC\_t_2^i,...,MC\_t_m^i} \}.
\end{equation}
The effect of encoding body parts with different colors for action ``right hand draw circle (clockwise)'' is illustrated in Fig.~\ref{fig3}, sub-figure (3). 

\subsubsection{Encoding Motion Magnitude}
\begin{figure}[!ht]
\begin{center}
{\includegraphics[height = 75mm, width = 80mm]{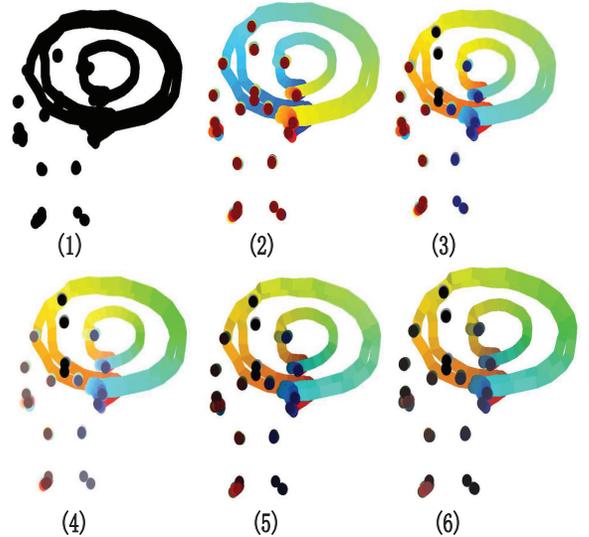}}
\end{center}
\caption{Step-by-step illustration of the front JTM for action ``right hand draw circle (clockwise)'' from the UTD-MHAD~\cite{chenchen2015b} dataset. (1) Joint trajectory map without encoding any motion direction and magnitude; (2) encoding joint motion direction in hue, where color variations indicate motion direction; (3) encoding body parts with different colormaps; (4) encoding motion magnitude into saturation; (5) encoding motion magnitude into brightness; (6) final JTM with all encodings.}
\label{fig3}
\end{figure}

Motion magnitude is one of the important factors in human motion. For an action, large magnitude of motion is likely to carry discriminative information. This paper proposes to encode the motion magnitude of joints into saturation and brightness so that the changes in motion would result in texture in the JMTs. Such texture is expected to be beneficial for ConvNets to learn discriminative features. For joints with high motion magnitude or speed, high saturation will be assigned. Specifically, the saturation is set to range from $s_{min}$ to $s_{max}$.  Given a trajectory, its saturation $S_{j}^{i}$ along the path of the trajectory could be calculated as
\begin{equation}
S_{j}^{i} = \frac{v_{j}^{i}}{max\{v\}} \times (s_{max} - s_{min}) + s_{min}
\label{eq:velocity}
\end{equation}
where $v_{j}^{i}$ is the speed of $j$th joint at the $i$th frame.
\begin{equation}
v_{j}^{i} = \Vert P_{j}^{i+1} - P_{j}^{i} \Vert_{2}
\label{eq:velocity}
\end{equation}
Let a trajectory modulated by saturation be $M{C_s}\_t_k^i$,  function $f(i)$ is refined as:
\begin{equation}
f(i) = \{{M{C_s}\_t_1^i,M{C_s}\_t_2^i,...,M{C_s}\_t_m^i} \}
\end{equation}
For the sample example in Fig.~\ref{fig3}, the encoding effect can be seen in the sub-figures (4), where the slow motion becomes diluted (e.g. trajectory of knees and ankles) while the fast motion becomes saturated (e.g. the green part of the circle).

To further enhance the motion patterns in the JTMs, the brightness is modulated by the speed of joints. Given a trajectory $t_{j}^{i}$ whose speed is $v_{j}^{i}$, its brightness $B_{j}^{i}$ is computed as
\begin{equation}
B_{j}^{i} = \frac{v_{j}^{i}}{max\{v\}} \times (b_{max} - b_{min}) + b_{min}
\label{eq:velocity1}
\end{equation}
where $b_{min}$ and $b_{max}$ represent the range of the brightness. Let $M{C_b}\_t_k^i$ be the trajectory with brightness and function $f(i)$ is then updated to:
\begin{equation}
f(i) = \{ {M{C_b}\_t_1^i,M{C_b}\_t_2^i,...,M{C_b}\_t_m^i} \}. 
\end{equation}
The effect of brightness modulation can be seen in sub-figure (5) of the example shown in in Fig.~\ref{fig3}, where texture becomes apparent (e.g. the yellow parts of the circle).

Finally, let $M{C_{sb}}\_t_k^i$ be the  trajectory after encoding the motion magnitude into both saturation and brightness. Function $f(i)$ can be expressed as:
\begin{equation}
f(i) = \{ {M{C_{sb}}\_t_1^i,M{C_{sb}}\_t_2^i,...,M{C_{sb}}\_t_m^i} \}.
\end{equation}
As illustrated in Fig.~\ref{fig3}, sub-figure (6), the motion variation enriches the texture in the final JTM.

\subsection{ConvNets Training}
After constructing the three JTMs on three orthogonal image planes, three ConvNets are fine-tuned individually, each ConvNet is an AlexNet~\cite{krizhevsky2012imagenet}. The fine-tuning procedure is similar to the one in~\cite{krizhevsky2012imagenet}. The network weights are learned using the mini-batch stochastic gradient descent with the momentum being set to 0.9 and weight decay being set to 0.0005. All hidden weight layers use the rectification (RELU) activation function. At each iteration, a mini-batch of 256 samples is constructed by sampling 256 shuffled training samples. The images are resized to 256 $\times$ 256. The learning rate is set to $10^{-2}$ for training from scratch and set to $10^{-3}$ for fine-tuning with pre-trained models on ILSVRC-2012, and then it is decreased according to a fixed schedule. For each ConvNet the training undergoes 100 cycles and the learning rate decreases every 30 cycles. For all experiments, the dropout regularization ratio was set to 0.9 in order to reduce complex co-adaptations of neurons in the nets for both networks.

\subsection{Multiply Score Fusion}
Given a testing skeleton sequence (sample), three JTMs are generated and fed into the three ConvNets respectively. Multiply score fusion is used to combine the outputs from the individual ConvNets. Specifically, the score vectors outputted by the three ConvNets are multiplied in an element-wise way, and the max score in the resultant vector is assigned as the probability of the test sequence. The index of this max score corresponds to the recognized class label.

\section{Experiments}\label{experiments}
The proposed method was evaluated on four public benchmark datasets: the large NTU RGB+D Dataset~\cite{shahroudy2016ntu}, MSRC-12 Kinect Gesture Dataset~\cite{Fothergill2011}, G3D~\cite{bloom2012g3d} and UTD-MHAD~\cite{chenchen2015b}. Experiments were conducted on the effectiveness of individual encoding scheme in the proposed method, the effectiveness of rotation, the role of fine-tuning, and the multiply score fusion compared with the max and average score fusion methods. The final recognition results were compared with the state-of-the-art reported on the same datasets. In all experiments, the saturation and brightness range from 0\% $\sim$ 100\% (mapped to 0 $\sim$ 255 in the JTM images).

\subsection{Evaluation of Key Design Factors}

\subsubsection{Different Encoding Schemes}

The effectiveness of different encoding schemes (as illustrated in Fig.~\ref{fig3}) was evaluated on the G3D dataset, and the recognition accuracies are listed in Table~\ref{steps}.

\begin{table}[!th]
\centering
\caption{Comparison of the Different Encoding Schemes on the G3D Dataset in terms of recognition accuracy.\label{steps}}
\begin{tabular}{|c|c|c|c|c|}
\hline
Techniques & Front &  Top &  Side & Fusion\\
\hline
Trajectory: $t_1^i$ & 65.45\% & 72.18\% & 73.54\% & 80.58\%\\
\hline
Trajectory: $C\_t_1^i$ & 76.12\%& 75.55\%& 76.56\% & 83.65\%\\
\hline
Trajectory: $MC\_t_1^i$ & 79.98\%& 78.25\%& 79.40\% & 87.68\%\\
\hline
Trajectory: $M{C_s}\_t_1^i$ & 83.52\%& 81.32\% &82.08\% & 89.98\% \\
\hline
Trajectory: $M{C_b}\_t_1^i$ & 84.46\% & 84.68\%& 85.60\%& 93.84\% \\
\hline
Trajectory: $M{C_{sb}}\_t_1^i$ & 86.25\%& 87.56\% &86.54\% & 96.02\% \\
\hline
\end{tabular}

\end{table}
From Table~\ref{steps} we can see that the proposed encoding methods effectively capture  spatio-temporal information. Each encoding method gradually amends more information to the JTMs for the three ConvNets to learn the discriminative features and improves the recognition. The three JTMs are complimentary to each other to improve recognition significantly through fusion.

\subsubsection{Rotation}

Rotation is adopted to mimic multiple views, and this simple process makes the proposed method capable of cross-view action recognition. At the same time, the rotation enlarges the training data and enables the method to work on small datasets. Table~\ref{rotation} shows the comparison of the proposed method with and without rotation on the NTU RGB+D and G3D datasets. As expected, the rotation operation improves the performance of cross-view recognition largely (by almost 3.5 percentage points).

\begin{table}[!th]
\centering
\caption{Comparison the proposed method with and without rotation on the NTU RGB+D and G3D datasets in Terms of Recognition Accuracy.\label{rotation}}
\begin{tabular}{|c|c|c|}
\hline
Dataset & \tabincell{c}{Without\\Rotation} &  \tabincell{c}{With\\Rotation} \\
\hline
NTU RGB+D (Cross Subject)  & 75.30\% & 76.32\%\\
\hline
NTU RGB+D (Cross View)  & 77.67\% & 81.08\%  \\
\hline
G3D & 95.12\%& 96.02\% \\
\hline
\end{tabular}
\end{table}

\subsubsection{Fine-tuning vs. Training from Scratch}

Even though the number of training samples per class is over 600 for the NTU RGB+D Dataset,  fine-tuning with available models from ImageNet is still preferred in terms of recognition accuracy. Table~\ref{finetuning} shows the results of two settings, fine-tuning and training from scratch, on NTU RGB+D and G3D datasets. In both settings, no rotation was performed. Notice that fine-tuning improved the recognition by 5 percentage point on the NTU RGB+D Dataset and almost doubled the recognition accuracy on the small G3D Dataset compared to training from scratch.

\begin{table}[!th]
\centering
\caption{Comparisons of fine-tuning and tranining from scratch on the NTU RGB+D and G3D datasets in Terms of Recognition Accuracy.\label{finetuning}}
\begin{tabular}{|c|c|c|}
\hline
Dataset &\tabincell{c}{Training from Scratch} &  \tabincell{c}{Fine-tuning} \\
\hline
NTU RGB+D (Cross Subject)  & 72.50\% & 75.30\%\\
\hline
NTU RGB+D (Cross View)  & 73.77\% & 77.67\%  \\
\hline
G3D  & 46.64\% & 94.65\% \\
\hline
\end{tabular}
\end{table}

\subsubsection{Comparison of Three Score Fusion Methods}

There are two common used late score fusion methods, namely, average score fusion method and max score fusion method. However, in this paper, we propose to adopt multiply score fusion which turns out to be more effective on the evaluated datasets. The comparison of these three score fusion methods on the four datasets for final recognition are listed in Table~\ref{fusion}. From the Table we can see that on the evaluated four datasets, the multiply score fusion consistently outperformed the average and max score fusion methods. This verifies that the three JTMs are likely to be statistically independent and provide complementary information.

\begin{table}[!th]
\centering
\caption{Comparison of Three Score Fusion Methods on the Four Datasets in Terms of Recognition Accuracy.\label{fusion}}
\begin{tabular}{|c|c|c|c|}
\hline
Dataset & Max &  Average &  Multiply \\
\hline
NTU RGB+D (Cross Subject) & 73.56\% & 75.05\% & 76.32\% \\
\hline
NTU RGB+D (Cross View) & 78.43\% & 79.88\% & 81.08\% \\
\hline
MSRC-12 & 91.70\%& 93.42\%& 94.86\% \\
\hline
G3D & 93.78\%& 94.65\%& 96.02\% \\
\hline
UTD-MHAD & 85.81\%& 86.42\% & 87.90\%  \\
\hline
\end{tabular}
\end{table}

\subsection{ NTU RGB+D Dataset}
 To our best knowledge, NTU RGB+D Dataset is currently the largest action recognition dataset. The 3D data is captured by Kinect v2 cameras. The dataset has more than 56 thousand sequences and 4 million frames, containing 60 actions performed by 40 subjects aging between 10 and 35. It consists of front view, two side views and left, right 45 degree views. This dataset is challenging due to large intra-class and viewpoint variations.
 
For fair comparison and evaluation, the same protocol as that in~\cite{shahroudy2016ntu} was used. It has both cross-subject and cross-view evaluation. In the cross-subject evaluation, samples of subjects 1, 2, 4, 5, 8, 9, 13, 14, 15, 16, 17, 18, 19, 25, 27, 28, 31, 34, 35 and 38 were used as training and samples of the remaining subjects were reserved for testing. In the cross-view evaluation, samples taken by cameras 2 and 3 were used as training, testing set includes the samples of camera 1. Table~\ref{tableNTU} lists the performance of the proposed method and those reported before.
 \begin{table}[h]\small
\setlength{\belowcaptionskip}{9pt}
  \centering
 \caption{Comparative Accuracies of the Proposed Method and Previous Methods on NTU RGB+D Dataset.\label{tableNTU} }
 \begin{tabular}{ccc}
  \hline
  Method&	Cross subject	&  Cross view\\
    \hline
Lie Group~\cite{vemulapalli2014human}	&50.08\%	 & 52.76\%\\

Dynamic Skeletons~\cite{ohn2013joint}	&60.23\%	 & 65.22\%\\

HBRNN~\cite{du2015hierarchical}	&59.07\%	 & 63.97\%\\

2 Layer RNN~\cite{shahroudy2016ntu}	&56.29\% &	64.09\%\\

2 Layer LSTM~\cite{shahroudy2016ntu}	&60.69\%	&67.29\%\\

Part-aware LSTM~\cite{shahroudy2016ntu}&	62.93\%	&70.27\%\\

ST-LSTM~\cite{liu2016spatio} &65.20\%	&76.10\%	\\
ST-LSTM+ Trust Gate~\cite{liu2016spatio}&	69.20\%	&77.70\%\\

Proposed Method	&\bfseries{76.32}\%&	\bfseries{81.08}\%\\
\hline

\end{tabular}
\end{table}
From this Table we can see that our proposed method achieved the state-of-the-art results compared with both hand-crafted features and deep learning methods.
The work~\cite{vemulapalli2014human} focused only on single person action and could not model multi-person interactions well. Dynamic Skeletons method~\cite{ohn2013joint} performed better than some RNN-based methods verifying the weakness of the RNNs~\cite{du2015hierarchical,shahroudy2016ntu}, which only mines the short-term dynamics and tends to overemphasize the temporal information even on large training data.  LSTM and its variants~\cite{shahroudy2016ntu,liu2016spatio} performed better due to their ability to utilize long-term context compared to conventional RNNs, but it is still weak in exploiting spatial information.  The proposed method achieved the best results in both cross-subject and cross-view evaluation.

\subsection{MSRC-12 Kinect Gesture Dataset}
MSRC-12~\cite{Fothergill2011} is a relatively large dataset for gesture/action recognition from 3D skeleton data captured by a Kinect sensor. The dataset has 594 sequences, containing 12 gestures by 30 subjects, 6244  gesture instances in total. The 12 gestures are: ``lift outstretched arms", ``duck", ``push right", ``goggles", ``wind it up", ``shoot",  ``bow", ``throw", ``had enough", ``beat both", ''change weapon" and ``kick". For this dataset, cross-subjects protocol was adopted, that is, odd subjects were used for training and even subjects were for testing. Table~\ref{table1} lists the performance of the proposed method and the results reported before.

\begin{table}[!th]
\centering
\caption{Comparison of the Proposed Method with the Existing Methods on the MSRC-12 Kinect Gesture Dataset.\label{table1}}
\begin{tabular}{|c|c|}
\hline
Method & Accuracy (\%)\\
\hline
HGM~\cite{yang2014hierarchical} & 66.25\%\\
\hline
Pose-Lexicon~\cite{zhou2016} & 85.86\%\\
\hline
ELC-KSVD~\cite{zhou2014discriminative} & 90.22\%\\
\hline
Cov3DJ~\cite{Hussein2013} & 91.70\%\\
\hline
SOS~\cite{Hou2016} & 94.27\%\\
\hline
Proposed Method & \textbf{94.86\%}  \\
\hline
\end{tabular}

\end{table}

\begin{figure}[!ht]
\begin{center}
{\includegraphics[height = 50mm, width = 85mm]{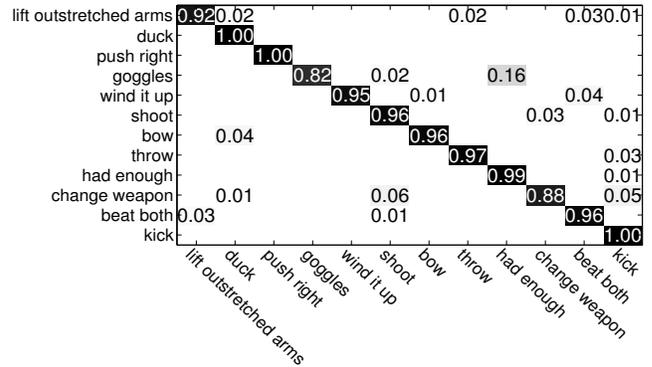}}
\end{center}
\caption{The confusion matrix of the proposed method on the MSRC-12 Kinect gesture dataset.}
\label{fig:confusion1}
\end{figure}

The confusion matrix is shown in Fig.~\ref{fig:confusion1}. From the confusion matrix we can see that the proposed method distinguishes most of actions very well, but it is not very effective to distinguish ``goggles" and ``had enough" which shares the similar appearance of JTMs probably caused by 3D to 2D projection.

\subsection{G3D Dataset}
Gaming 3D Dataset (G3D)~\cite{bloom2012g3d} focuses on real-time action recognition in a gaming scenario. It contains 10 subjects performing 20 gaming actions: ``punch right", ``punch left", ``kick right", ``kick left", ``defend", ``golf swing", ``tennis swing forehand", ``tennis swing backhand", ``tennis serve", ``throw bowling ball", ``aim and fire gun", ``walk", ``run", ``jump", ``climb", ``crouch", ``steer a car", ``wave", ``flap" and ``clap".
For this dataset, the first 4 subjects were used for training, the fifth for validation and the remaining 5 subjects were for testing as configured in~\cite{Nie2015}. Table~\ref{table2} compared the performance of the proposed method and those reported in~\cite{Nie2015}.

\begin{table}[ht!]
\centering
\caption{Comparison of the Proposed Method with Previous Methods on  the G3D Dataset.\label{table2}}
\begin{tabular}{|c|c|}
\hline
Method & Accuracy (\%)\\
\hline
Cov3DJ~\cite{Hussein2013} & 71.95\%\\
\hline
ELC-KSVD~\cite{zhou2014discriminative} & 82.37\%\\
\hline
LRBM~\cite{Nie2015} & 90.50\%  \\
\hline
SOS~\cite{Hou2016} & 95.45\%\\
\hline
Proposed Method & \textbf{96.02\%}\\
\hline
\end{tabular}

\end{table}

\begin{figure}[!ht]
\begin{center}
{\includegraphics[height = 70mm, width = 85mm]{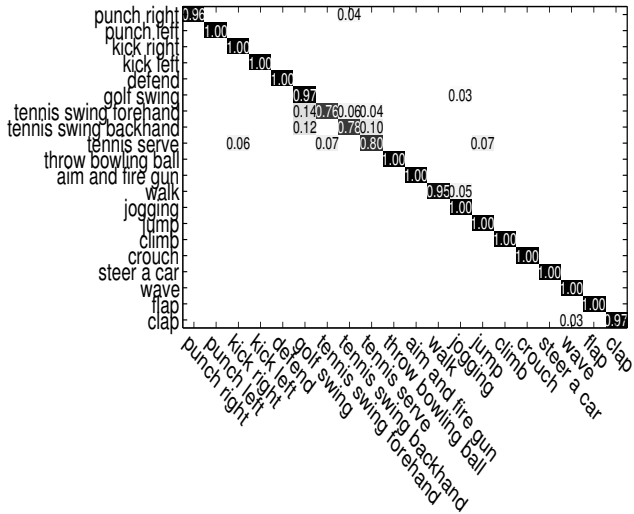}}
\end{center}
\caption{The confusion matrix of the proposed method on the G3D Dataset.}
\label{fig:confusion2}
\end{figure}

The confusion matrix is shown in figure~\ref{fig:confusion2}. From the confusion matrix we can see that the proposed method recognizes most of actions well. The proposed method outperformed LRBM. LRBM confused the actions among ``tennis swing forehand" and ``bowling", ``golf" and ``tennis swing backhand", ``aim and fire gun" and ``wave", ``jump" and ``walk", however, these actions are well distinguished by the proposed method likely because of the quality spatial information encoded in the JTMs. As for ``aim and fire gun" and ``wave", the proposed method could not distinguish them well without encoding the motion magnitude, but does well with the encoding of motion magnitude. However, the proposed method, confused ``tennis swing forehand" and ``tennis swing backhand". It's probably because the front and side projections of body shape of the two actions are too similar.

\subsection{UTD-MHAD}

UTD-MHAD~\cite{chenchen2015b} is a multimodal action dataset, captured by one Microsoft Kinect camera and one wearable inertial sensor. This dataset contains 27 actions performed by 8 subjects (4 females and 4 males) with each subject performing each action 4 times. After removing three corrupted sequences, the dataset has 861 sequences. The actions are: ``right arm swipe to the left", ``right arm swipe to the right", ``right hand wave", ``two hand front clap", ``right arm throw", ``cross arms in the chest", ``basketball shoot", ``right hand draw x", ``right hand draw circle (clockwise)", ``right hand draw circle (counter clockwise)", ``draw triangle", ``bowling (right hand)", ``front boxing", ``baseball swing from right", ``tennis right hand forehand swing", ``arm curl (two arms)", ``tennis serve", ``two hand push", ``right hand know on door", ``right hand catch an object", ``right hand pick up and throw", ``jogging in place", ``walking in place", ``sit to stand", ``stand to sit", ``forward lunge (left foot forward)" and ``squat (two arms stretch out)". It covers sport actions (e.g. ``bowling", ``tennis serve" and ``baseball swing"), hand gestures (e.g. ``draw X", ``draw triangle", and ``draw circle"), daily activities (e.g. ``knock on door", ``sit to stand" and ``stand to sit") and training exercises (e.g. ``arm curl", ``lung" and ``squat"). For this dataset, cross-subjects protocol was adopted as in ~\cite{chenchen2015b}, namely, the
data from the subjects numbered 1, 3, 5, 7 were used for training while subjects 2, 4, 6, 8 
were used for testing. Table~\ref{table3} compares the performance of the proposed method and those reported in~\cite{chenchen2015b}.

\begin{table}[ht!]
\centering
\caption{Comparison of the Proposed Method with the Previous Methods on UTD-MHAD Dataset.\label{table3}}
\begin{tabular}{|c|c|}
\hline
Method & Accuracy (\%)\\
\hline
ELC-KSVD~\cite{zhou2014discriminative} & 76.19\%\\
\hline
Kinect \& Inertial \cite{chenchen2015b} & 79.10\%  \\
\hline
Cov3DJ~\cite{Hussein2013} & 85.58\%\\
\hline
SOS~\cite{Hou2016} & 86.97\%\\
\hline
Proposed Method & \textbf{87.90\%}\\
\hline
\end{tabular}

\end{table}
Please notice that the method used in~\cite{chenchen2015b} is based on Depth and Inertial sensor data, not skeleton data alone.

\begin{figure}[!ht]
\begin{center}
{\includegraphics[height = 70mm, width = 85mm]{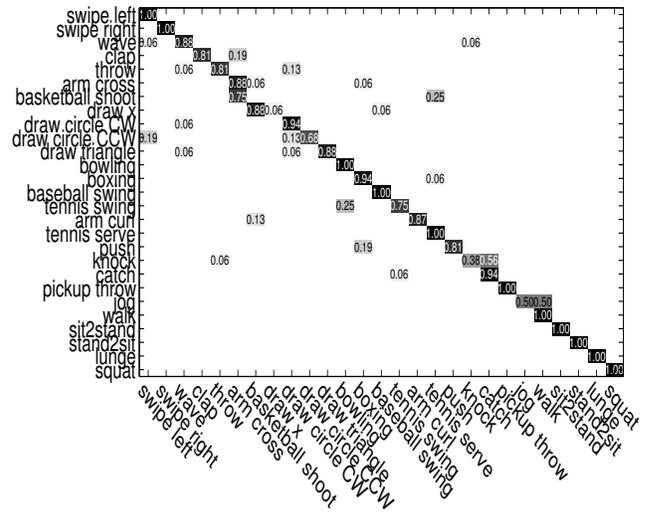}}
\end{center}
\caption{The confusion matrix of the proposed method on the UTD-MHAD dataset.}
\label{fig:confusion3}
\end{figure}

The confusion matrix is shown in Fig.~\ref{fig:confusion3}. This dataset is much more challenging compared to the previous two datasets. From the confusion matrix we can see that the proposed method can not distinguish some actions well, for example, ``jog" and ``walk". A probable reason is that the proposed encoding process is also a temporal normalization process. The actions ``jog" and ``walk" would be normalized to have similar JTMs after the encoding.

\subsection{Discussion}
In this paper, we adopt the three orthogonal planes of the natural real coordinates of the camera. One question is whether there are some orthogonal planes better than the natural ones. Generally speaking, there possibly exist three orthogonal views which are better than the natural coordinates if the three views result in less self-occlusion among the joints for all actions. Since only very sparse 20 joints are used to represent the skeleton, the likelihood of such self-occlusion of the joints would be very small. Consequently, no particular three orthogonal views would be obviously superior to others. However, the depth camera only captures $2\frac{1}{2}D$ in the natural coordinates and the skeleton is estimated from the $2\frac{1}{2}D$. It is likely that the natural coordinates could be slightly, but not significantly, better than other three orthogonal views. 

To validate this, we conducted the following experiments on the G3D Dataset. Different three orthogonal views were generated by rotating the 3D points of joints and projecting them to the three orthogonal planes. The rotation was performed with a fixed step of $22.5^{\circ}$ along the polar angle $\theta$ and azimuthal angle $\psi$, both in the range of  $[-45^{\circ}, 45^{\circ}]$. Note that this range effectively covers all possible views since rotation beyond this range would result in swapping of views. Such swapping would not affect the recognition accuracy after fusion. Table~\ref{Orthognal} shows the recognition accuracies of different orthogonal views indicated by the values of $\theta$ and $\psi$.

\begin{table}[!ht]
\centering

\caption{The Recognition Accuracy (\%) of Different Orthognal Views.\label{Orthognal}}
\begin{tabular}{|c|c|c|c|c|c|}
 \hline
 \diagbox{$\psi$}{$\theta$} & $~-45^{\circ}$ & $-22.5^{\circ}$ & $~~~0^{\circ}~~~$ &$~22.5^{\circ}~$ &$~~45^{\circ}~$ \\
 \hline
$-45^{\circ}$ & 94.45 & 92.12 & 94.85 & 92.24 & 92.42\\
\hline
$-22.5^{\circ}$ & 95.40  & 94.45 & 94.45 & 92.73 & 92.24\\
\hline
$0^{\circ}$& 94.45 & 95.05  & 95.12 & 94.85 & 94.15  \\
\hline
$22.5^{\circ}$ & 94.85 & 94.85 & 94.45 & 94.85 & 93.45\\
\hline
$45^{\circ}$ & 92.24 & 95.00 & 94.85 & 94.15 & 94.15\\
\hline
 \end{tabular}
\end{table}
The results in Table~\ref{Orthognal} have shown small and insignificant variation of the recognition accuracy among the views and the natural coordinates produced the best result.

In this paper, we fuse three orthogonal image planes to improve the final accuracy. Another questions is whether adding more views will lead to better recognition.  Some experiments were conducted on the G3D dataset to answer this question. Firstly, the views of the natural coordinates were fused with the views after rotating the points by the specified angles in $\theta$ and $\psi$. Table~\ref{Plusing} shows the results by fusing two pairs of three orthogonal planes, one is the natural coordinates and the other is specified by the rotation angles $\theta$ and $\psi$. The accuracies of all cases are almost same.

\begin{table}[!ht]
\centering
\caption{The Results of Fusing the Original Three Orthogonal Planes and Rotated Three Planes.\label{Plusing}}
\begin{tabular}{|c|c|c|c|c|c|}
 \hline
 \diagbox{$\psi$}{$\theta$} & $~-45^{\circ}$ & $-22.5^{\circ}$ & $~~~0^{\circ}~~~$ &$~22.5^{\circ}~$ &$~~45^{\circ}~$ \\
 \hline
$-45^{\circ}$ & 94.45 & 93.45 & 94.85 & 95.15 & 95.12\\
\hline
$-22.5^{\circ}$ & 94.85  & 94.54 & 95.15 & 95.12 & 94.85\\
\hline
$0^{\circ}$& 95.12 & 95.15  & 95.12 & 94.54 & 94.54  \\
\hline
$22.5^{\circ}$ & 94.85 & 94.24 & 95.15 & 95.15 & 94.85\\
\hline
$45^{\circ}$ & 94.85 & 94.85 & 95.15 & 95.12 & 95.15\\
\hline
\end{tabular}
\end{table}

We also evaluated the performance by fusing all views of the $9$ coordinates including the natural ones, where $\theta \in \{-22.5^{\circ}, 0^{\circ}, 22.5^{\circ}\}$ and $\psi \in \{-22.5^{\circ}, 0^{\circ}, 22.5^{\circ}\}$,  and all views of the $25$  coordinates, where $\theta \in \{-45^{\circ}, -22.5^{\circ},0^{\circ}, 22.5^{\circ},45^{\circ}\}$ and $\psi \in \{-45^{\circ}, -22.5^{\circ}, 0^{\circ}, 22.5^{\circ},45^{\circ}\}$ respectively. The results are shown in Table~\ref{FusingViews}. It can be seen that fusing views of multiple orthogonal coordinates did not improve the performance on this dataset. Similar results would be expected on other datasets for the reason explained above.

The above analysis and experiments have demonstrated that the three orthogonal views in the natural coordinates are likely to be sufficient. 

\begin{table}[!ht]
\centering
\caption{The Results for Fusing Views of Multiple Coordinates.\label{FusingViews}}
\begin{tabular}{|c|c|}
 \hline
 Number of Coordinates &  Accuracy (\%)  \\
\hline
9 & 95.15 \\
\hline
25 & 94.85 \\
\hline
\end{tabular}
\end{table}

\section{Conclusion}\label{conclusion}
This paper addresses the problem of human action recognition by applying ConvNets to skeleton sequences. An effective method is proposed to project the joint trajectories to three orthogonal JTMs to encode the spatial-temporal information into texture patterns. The three JTMs are complementary to each other.  Such image-based representation enables us to fine-tune the existing ConvNets models trained on image data for classification
of skeleton sequences, without  training the deep ConvNets afresh. The experimental results on the four datasets have shown the efficacy of the proposed encoding scheme. Extension of the proposed method to on-line action recognition is the focus of future work.

\ifCLASSOPTIONcaptionsoff
  \newpage
\fi



%





\begin{IEEEbiography}[{\includegraphics[width=1in,height=1.25in,clip,keepaspectratio]{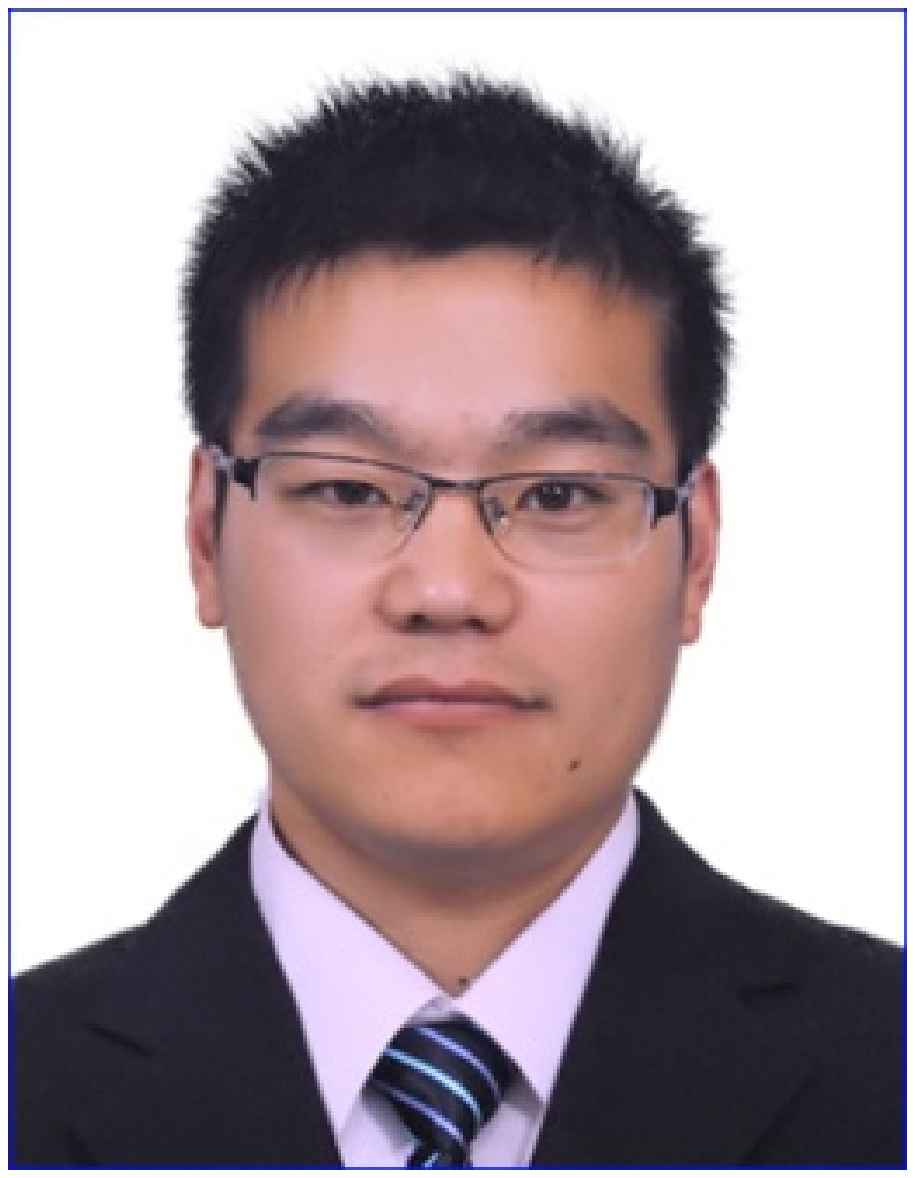}}]{Pichao Wang}

(S'14) received the BE degree in network engineering from Nanchang University, 
Nanchang, China, in 2010, and received the MS degree in communication and 
information system from Tianjin University, Tianjin, China, in 2013. He is 
currently pursuing the PhD degree with the School of Computing and Information Technology, University of Wollongong, Australia.
His current research 
interests include computer vision and machine learning. 
\end{IEEEbiography}
\begin{IEEEbiography}[{\includegraphics[width=1in,height=1.25in,clip,keepaspectratio]{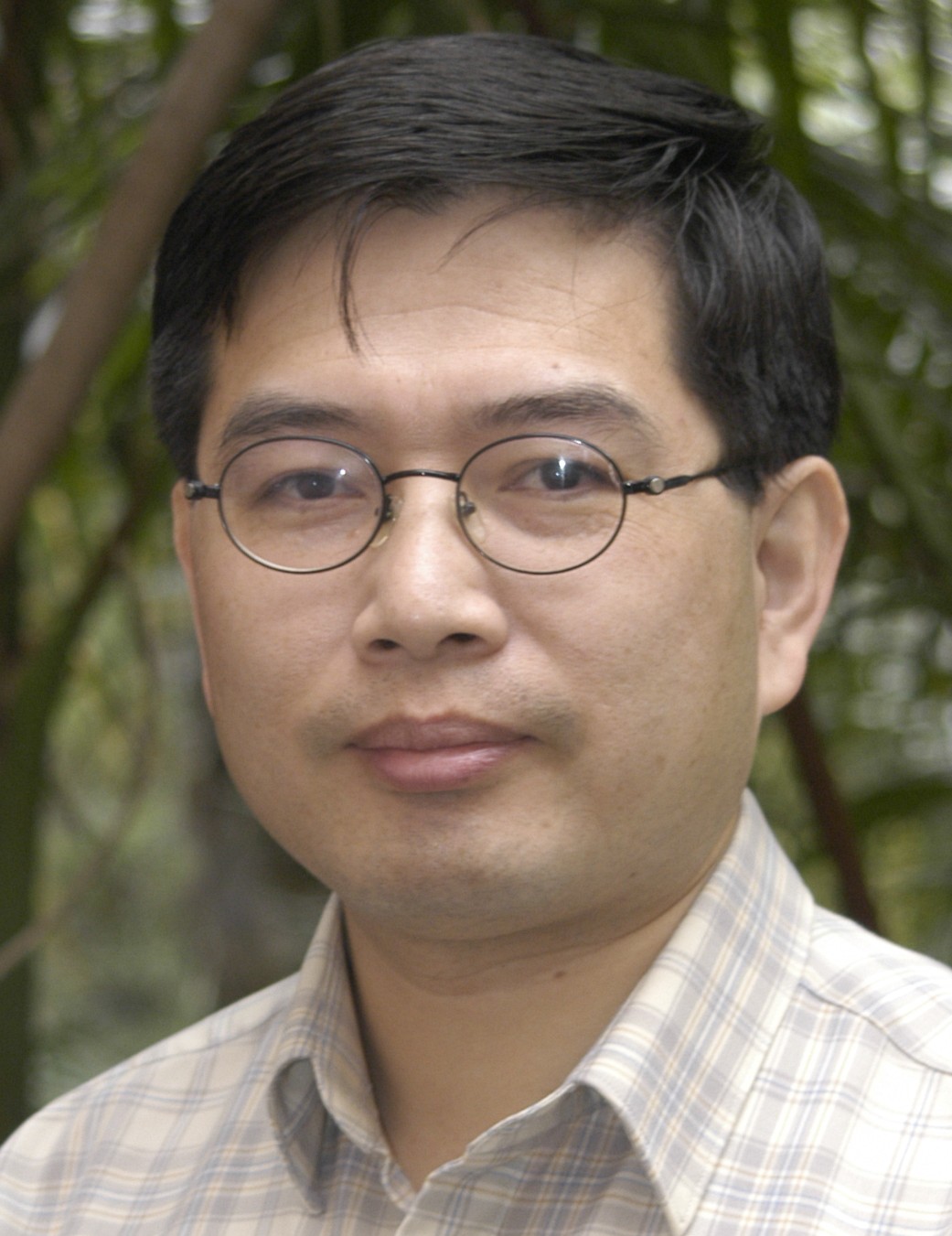}}]{Wanqing Li}
received his PhD in electronic engineering from The University of Western Australia. He is an Associate Professor and Co-Director of Advanced Multimedia Research Lab (AMRL) of University of Wollongong, Australia. His research areas are 3D computer vision, 3D multimedia signal processing and medical image analysis. Dr. Li is a Senior Member of IEEE.

\end{IEEEbiography}

\begin{IEEEbiography}[{\includegraphics[width=1in,height=1.25in,clip,keepaspectratio]{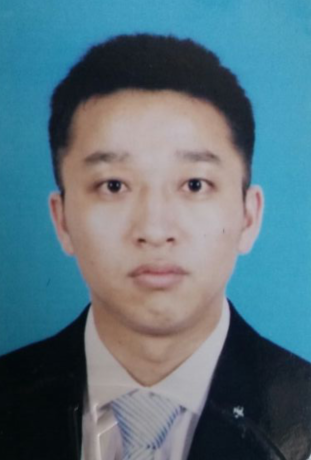}}]{Chuankun Li}

received the BE degree in electronic information engineering from North University of China , Taiyuan, China, in 2012 and received the MS degree in communication and information system from North University of China, Taiyuan, China, in 2015. He is currently pursuing the Ph.D degree with School of electronic information engineering , Tianjin University, China. His current research interests include computer vision and machine learning.
\end{IEEEbiography}

%

%
%

\begin{IEEEbiography}[{\includegraphics[width=1in,height=1.25in,clip,keepaspectratio]{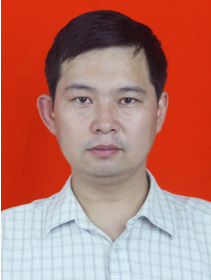}}]{Yonghong Hou}

(M'11) received the B.Eng. degree in electronic engineering from Xidian University, Xi’an,
China, in 1991, and the M.Eng. and Ph.D degrees both in communication and information system from
Tianjin University, Tianjin, China, in 2003 and 2009, respectively. Since 2006, he has been an associate professor of school of electronic and information engineering, Tianjin University. His
research interests include computer vision, artificial intelligence and multimedia signal processing
\end{IEEEbiography}

%

\end{document}